\documentclass[conference]{IEEEtran}
\IEEEoverridecommandlockouts

\usepackage{cite}
\usepackage{amsmath,amssymb,amsfonts}

\usepackage{algorithmic}
\usepackage{graphicx}
\usepackage{textcomp}
\usepackage[dvipsnames]{xcolor}
\usepackage{url}
\usepackage{soul}
\usepackage{xspace}
\usepackage{ulem}
\usepackage{multirow,tabularx}

\def\BibTeX{{\rm B\kern-.05em{\sc i\kern-.025em b}\kern-.08em
    T\kern-.1667em\lower.7ex\hbox{E}\kern-.125emX}}

\newcommand{\system}[1]{\textsc{#1}\xspace}
\newcommand{\bertlstm}{\system{BERT-LSTM}\xspace}
\newcommand{\seqseq}{\system{Seq2Seq}\xspace}

\title{Paraphrasing Techniques for Maritime QA system}

\author{\IEEEauthorblockN
{Fatemeh Shiri\textsuperscript{\textsection}, 
Terry Yue Zhuo\textsuperscript{\textsection}, Zhuang Li\textsuperscript{\textsection},\\
Shirui Pan, Weiqing Wang, Reza Haffari, Yuan-Fang Li } 
\IEEEauthorblockA{\textit{Department of Data Science and AI} \\
\textit{Faculty of Information Technology} \\
\textit{Monash University}, Melbourne, Australia \\
firstname.lastname@monash.edu\vspace{-20pt}\vspace{-5pt}}
\and
\IEEEauthorblockN{Van Nguyen}
 \IEEEauthorblockA{
\textit{Defence Science and Technology Group}\\
Adelaide, Australia \\
Van.Nguyen5@dst.defence.gov.au}}

\begin{document}
\maketitle
\begingroup\renewcommand\thefootnote{\textsection}
\footnotetext{Equal contribution}
\endgroup
\begin{abstract}

There has been an increasing interest in incorporating Artificial Intelligence (AI) into Defence and military systems to complement and augment human intelligence and capabilities.  However, much work still needs to be done toward achieving an effective human-machine partnership. This work is aimed at enhancing human-machine communications by developing a capability for automatically translating human natural language into a machine-understandable language (e.g., SQL queries). Techniques toward achieving this goal typically involve building a \emph{semantic parser} trained on a very large  amount of high-quality manually-annotated data. However, in many real-world Defense scenarios, it is not feasible to obtain such a large amount of training data. To the best of our knowledge, there are few works trying to explore the possibility of training a semantic parser with limited manually-paraphrased data, in other words, zero-shot. 
In this paper, we investigate how to exploit paraphrasing methods for the automated generation of large-scale training datasets (in the form of paraphrased utterances and their corresponding logical forms in SQL format) and present our experimental results using real-world data in the maritime domain.

\end{abstract}

\section{Introduction}
In recent years, there has been an increasing interest in incorporating 
Artificial Intelligence (AI) into Defence and military systems to complement and augment human intelligence and capabilities. For instance, AI technologies for Intelligent, Surveillance and Reconnaissance (ISR) now play a significant role in maintaining situational awareness and assist human partners with decision making. While AI technologies are outstanding at handling the enormous volumes of data, pattern recognition and anomaly detection, humans excel at making decisions on limited data and sense when data has been compromised (as witnessed with adversarial machine learning). A collaboration between machines and humans therefore holds great promise to help increasing adaptability, flexibility and performance across many Defence contexts. However much work still needs to be done toward achieving an effective human-machine partnership, including addressing the bottlenecks of communication, comprehension and trust. In this work, we are focused on the communication aspect which is at the center of human-machine coordination and success. 

Traditional methods require humans to interact with a machine using a machine language or controlled natural language (CNL) (e.g., \cite{saulwick2014lexpresso}). However, there is often no time for personnel to translate input to a machine and interpret complex machine outputs in military operations.
In our previous work, we proposed Maritime DeepDive \cite{shiri2021toward}, an automated construction of knowledge graphs from unstructured natural language data sources as part of the RUSH project \cite{nguyen2020fuzzy} for situational awareness. We are interested in endowing our situational awareness system with a capability to assist human decision makers by allowing them to interact with the system using human natural language. 

In this work, we investigate the application of state-of-the-art techniques in AI and natural language processing (NLP) for the automated translation of human natural language questions into SQL queries, toward enhancing human-machine partnership.  
Specifically, we build a semantic parser in the maritime domain starting with zero manually annotated training examples. Moreover, with the help of paraphrasing techniques, we reduce the gap between synthesized utterances and the natural real-world utterances.
Our  contributions are, 
\begin{itemize}
\item An efficient and effective approach for generating large-scale domain specific training dataset for Question Answering (QA) systems. Our training dataset includes synthetic samples and paraphrased samples obtained from state-of-the-art (SOTA) techniques. Therefore, the parser trained on such samples does not suffer from fundamental mismatches between the distributions of the automatically generated examples and the natural ones issued by real users.

\item Evaluation and analyses of various paraphrasing techniques to produce diverse natural language questions.

\item Discussion of lessons learned and recommendations.
\end{itemize}

\section{Background}
\subsection{Semantic Parsing}
Semantic parsing is a task of translating natural language utterances into formal meaning representations, such as SQL and abstract meaning representations (AMR)~\cite{banarescu2013abstract}. Modern neural network approaches usually formulate semantic parsing as a machine translation problem and model the semantic parsing process with variations of Seq2Seq~\cite{sutskever2014sequence} frameworks. The output of the Seq2Seq models are either linearized LF (Logic Form) token sequences~\cite{dong2016language,dong2018coarse,xu2020schema2qa} or action sequences that construct LFs~\cite{chen2018sequence,guo2019irnet,wang2019rat,rabinovich2017abstract,li2021few,li2021total}. Semantic parsing has a wide range of applications including question answering~\cite{chen2018sequence,guo2019irnet,wang2019rat}, programming synthesis~\cite{rabinovich2017abstract}, and natural language understanding (NLU) in dialogue systems~\cite{gupta2018semantic}.

Our work is mainly focused on question answering. The user questions expressed in natural language are converted into SQL queries which are then executed on our Maritime DeepDive Knowledge Graph~\cite{shiri2021toward} to retrieve answers to the questions. 
The recent surveys~\cite{Kamath2018semanticParsingSurvey,zhu2019survey,li2020context} cover comprehensive reviews about recent semantic parsing studies. Of particular relevance to this work are \textit{BootStrapping Semantic Parsers}.

\subsection{Bootstrapping Semantic Parsers}
Data scarcity is always a serious problem in semantic parsing since it is difficult for the annotators to acquire expert knowledge about the meaning of the target representations. To solve this problem, one line of research is to bootstrap semantic parsers with semi-automatic data synthesis methods. \cite{wang2015building, xu2020schema2qa} use a set of synchronous context-free grammar (SCFG) rules and canonical templates to generate a large number of \textit{clunky} utterance-SQL pairs, respectively. And then they hire crowd-workers to paraphrase the \textit{clunky} utterances into natural questions. In order to reduce the paraphrase cost, \cite{herzig2019don} applies a paraphrase detection to automatically align the clunky utterances with the user query logs. This method requires access to user query logs, which is infeasible in many scenarios. In addition, it requires human effort to filter out the false alignments. \cite{xu2020autoqa,yin2021ingredients} use automatic paraphrase models (e.g. fine-tuned BART~\cite{lewis2020bart}) to paraphrase clunky utterances and an automatic paraphrase filtering method to filter out low-quality paraphrases. The data synthesis method from \cite{xu2020autoqa,yin2021ingredients} requires the lowest paraphrase cost among all the aforementioned approaches. \cite{oren-etal-2021-finding} generates synthetic data using SCFGs as well. \cite{oren-etal-2021-finding} applies various approaches to down-sample a subset from the synthetic data. With their sampling method, training the parser with 200x less training data can perform comparably with training on the total population.



\section{Our proposed approach}\label{sec:approach}

\begin{figure}
    \centering
    \resizebox{.48\textwidth}{!}{
    \includegraphics{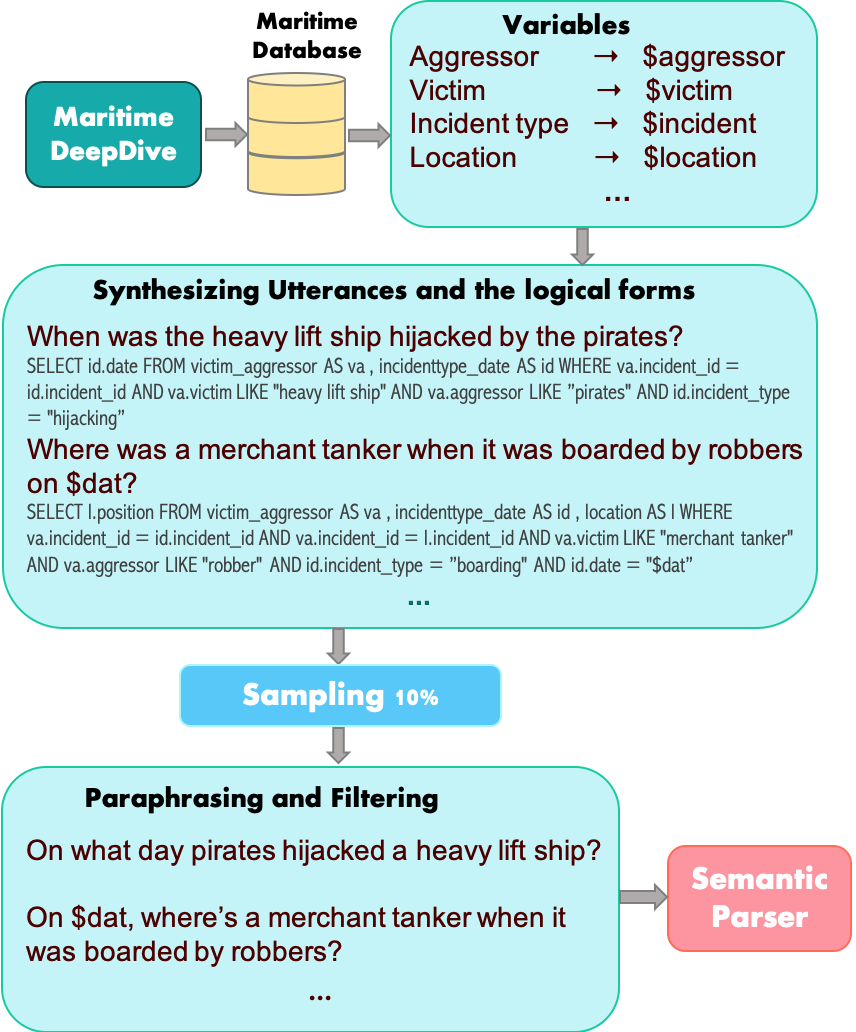}}
    \caption{An illustration of our end-to-end framework.}
    \label{fig:pipline}
\end{figure}

Our proposed approach, as shown in Figure \ref{fig:pipline}, improves on existing work (such as SEQ2SEQ and RoBERTa-based semantic parsers) for bootstrapping semantic parsers by not requiring manually annotated training data. Instead, large-scale training datasets can be generated in an automated manner through the use of existing automatic paraphrasing techniques. Specifically, we design a compact set of \textit{synchronous grammar} rules to generate \textit{seed} examples as pairs of canonical utterances and corresponding logical forms. We then apply a number of paraphrasing and filtering techniques to this initial set to create a much larger set of more diverse alternatives. 

\subsection{Semantic Parser}

We adopt two common methods for training semantic parsers: i) the attentional \seqseq~\cite{luong2015effective} framework which uses LSTMs~\cite{hochreiter1997long} as the encoder and decoder, and ii) \bertlstm~\cite{xu2020schema2qa}, a Seq2Seq framework with a copy mechanism~\cite{gu2016incorporating} which uses RoBERTa~\cite{liu2019roberta} as the encoder and LSTM as the decoder. The input to the Seq2Seq model is the natural-language question and  output is a sequence of linearized logical form's tokens.

\subsection{Synchronous grammar}
Our constructed Maritime DeepDive Knowledge Graph \cite{shiri2021toward}, containing a set of entities such as {\it{victim}}, {\it{aggressor}} and triples such as {\it{($e_1$, r, $e_2$)}}, where  {\it{$e_1$}} and  {\it{$e_2$}} are entities and {\it{r}} is a relationship (e.g.,  {\it{victim\_aggressor}}). The database can be queried using SQL logical forms.

\begin{figure*}
    \centering
    \resizebox{.9\textwidth}{!}{
    \includegraphics{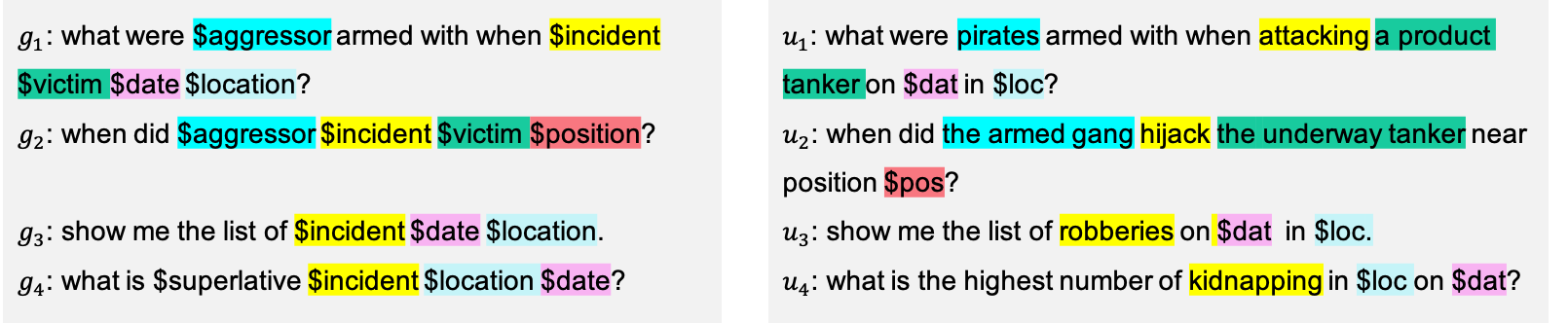}}
    \caption{Examples of utterances (right) generated from Synchronous grammar rules (left).}
    \label{fig:grammar}
\end{figure*}

We propose to design compact grammars with only 31 SCFG rules that simultaneously generate both logical forms and canonical utterances such that the utterances are understandable by a human. First, we define variables specifying a canonical phrase such as {\it{victim}},  {\it{aggressor}}, {\it{incident\_type}},  {\it{date}}, {\it{position}},  {\it{location}}. Second, we develop grammar rules for different SQL logical form structures (Figure \ref{fig:grammar}). Finally, our framework uses the grammar rules and the list of variables and their possible values ($G$ , $L$) to automatically generate canonical utterances paired with their SQL logical forms ($u$ , $lf$) exhaustively (Figure \ref{fig:grammar}). It yields a large set of canonical examples to train a semantic parser. We assign real values to the domain-specific variables including {\it{victim}} (victimized ships and individuals, e.g. oil tanker),  {\it{aggressor}} (e.g. pirates), {\it{incident\_type}} (e.g. robbery and hijacking). However, we replace all the general {\it{location}} (approximate place, e.g. country), {\it{position}} (place with longitude and latitude) and {\it{date}} with abstract variables  $\$loc$,  $\$pos$ and $\$dat$, which later can be used to capture the actual content using a NER model and parse into SQL queries. The location, position and date variables can have various unlimited content, and we do not restrict the parser to some content.  

\subsection{Paraphrase Generation}
\begin{table*}[t!]
\caption{Examples of paraphrasing using different techniques.}
\vspace{-0.1 cm}
\centering
\resizebox{\textwidth}{!}{
\begin{tabular}{ll}
\hline
\textbf{Canonical   Utterance:}          & which weapon did pirates use to rob the offshore supply vessel on $\$dat$ in $\$loc$?       \\ \hline
\multicolumn{1}{l|}{\textbf{Techniques}} & \textbf{Paraphrased  Utterances}                              \\ \hline
\multicolumn{1}{l|}{Back-translation (Spanish)}    &  what weapon did the pirates used to steal the offshore supply ship in $\$dat$ at $\$loc$?   \\
\multicolumn{1}{l|}{Back-translation (Telugu)}    &  which weapon is used to rob the offshore supply vessel in $\$dat$ by pirates in $\$loc$?  \\
\multicolumn{1}{l|}{Back-translation (Chinese)}    &  which weapon is used in $\$loc$? it is used to rob the offshore supply boat? \\
\multicolumn{1}{l|}{GPT-3}               & what was the weapon used by pirates to rob the offshore supply vessel on $\$dat$ in $\$loc?$ \\
\multicolumn{1}{l|}{BART}                & what gun did the pirates use to rob an offshore supply vessel in $\$loc$ on $\$dat$? \\
\multicolumn{1}{l|}{Quillbot}            & when pirates robbed an offshore supply vessel on $\$dat$ in $\$loc$, what weapon did they use?                         \\ \hline
\end{tabular}
\centering
}
\label{tab:examples}
\end{table*}
The paraphrase generation model rewrites the canonical utterance to more diverse alternatives, which later are used to train the semantic parser. Existing work do not fully explore current paraphrasing approaches. Instead, they resort to applying language models (e.g., BART and GPT-3) to paraphrase utterances in datasets and do not tailor to specific domains.

We improve on the existing work and explore other augmentation methods for \textit{paraphrase generation} on domain-specific data, making use of \textbf{back-translation}, \textbf{prompt-based language models}, \textbf{fine-tuned autoregressive language model} and \textbf{commercial paraphrasing model}.

\subsubsection{Paraphrasing using back-translation}
Back Translation is a data augmentation and evaluation technique that is widely used in several studies \cite{beddiar2021data,romaissa2021data, prabhumoye2018style,zhuo2022rethinking}. Given a sentence, we aim to translate it to another language and translate it back, where the back-translated sentence will be slightly different from the original one. We thoroughly compared the performance of Google Translation API, a well-known translation tool for daily use. Google Translate API can translate 109 languages in all. It is supported by powerful neural machine translation methods enhanced with advanced techniques in a well-designed pipeline architecture.

To ensure the diversity of paraphrased data, we adopt the procedure as defined in \cite{corbeil2020bet}:
\begin{itemize}
    \item Clustering languages into different family branches via Wikipedia info-boxes.
    \item Selecting the most used languages from each family and translating them using an appropriate translation system.
    
    \item Keeping the top three languages with the best translation performance.
    
\end{itemize}

\subsubsection{Paraphrasing using prompt-based language model generation}
Several studies \cite{shin2021few, shin2021constrained, schucher2021power} have investigated few-shot large pre-trained language models such as GPT-3 \cite{brown2020language} and ChatGPT~\cite{zhuo2023exploring} for semantic parsing. Unlike the `traditional' approach where a pre-trained language model can be leveraged by adapting its parameters to the task at hand through fine-tuning (\textit{language model as pre-training task}), GPT-3 took a different approach where it can be treated as a ready-to-use multi-task solver (\textit{language modelling as multi-task learning}). This could be achieved by transforming certain tasks we want it to solve into the form of language modelling. Specifically, by designing and constructing an appropriate input sequence of words (called a \textit{prompt}), one is able to induce the model to produce the desired output sequence (i.e., a paraphrased utterance in this context)without changing its parameters through fine-tuning at all. In particular, in the low-data regime, empirical analysis shows that, either for manually picking hand-crafted prompts [21] or automatically building auto-generated prompts [8], [16] taking prompts for tuning models is surprisingly effective for the knowledge stimulation and model adaptation of pre-trained language model. However, none of these methods reports the use of a few-shot pre-trained language model to directly generate few-shot and zero-shot paraphrased utterances. In this work, we supply GPT-3 with an instructive prompt to generate paraphrases in a zero-shot setting.   

\subsubsection{Paraphrasing using fine-tuned language model generation}
Previous researches have proposed several pretrained autoregressive language models, such as GPT-2 \cite{radford2019language} and ProphetNet \cite{qi2020prophetnet}. Other studies such as \cite{moghimifar2020cosmo} use these language models to solve various downstream tasks and achieve SOTA performance. In this work, following~\cite{yin2021ingredients}, we fine-tune an autoregressive language model BART~\cite{lewis2020bart} on the dataset from~\cite{krishna2020reformulating}, which is a subset of the PARANMT corpus~\cite{wieting2018paranmt}, to generate syntactically and lexically diversified paraphrases.
\subsubsection{Paraphrasing using a commercial system} 
We also experiment with paraphrase generation using Quillbot.\footnote{https://quillbot.com} Quillbot is a commercial tool that provides a scalable and robust paraphraser that can control synonyms and generation styles. 
\subsection{Paraphrase Filtering}
\label{sec:para_filter}
Since the automatically generated paraphrases may have varying quality, we further filter out the paraphrases of low quality. In our work, we adopt the filtering method discussed in \cite{xu2020autoqa} in the spirit of self-training. The process consists of the following steps:
\begin{enumerate}
    \item Evaluate the parser on the generated paraphrases and keep those for which the corresponding SQL logical forms are correctly generated by the parser.
    \item Add the paraphrase-SQL pairs into the training data and re-train the parser.
    \item Repeat steps 1--2 for several rounds or until no more paraphrases are kept. 
\end{enumerate}

This method is based on three assumptions~\cite{xu2020autoqa}: i) the parser could generalize well to unseen paraphrases which share the same semantics with the original questions, ii) the synthetic dataset generated by the SCFGs are good enough to train an initial model, and iii) it is very unlikely for a poor parser to generate correct SQL queries by chance. To improve the model's generalization ability, we use \bertlstm instead of vanilla Seq2Seq as our base parser for filtering.

In the following section, we provide an experimental analysis of the performance of our proposed approach.


\section{Experiments}
\subsection{Maritime DeepDive}
We utilize Maritime DeepDive \cite{shiri2021toward}, a probabilistic knowledge graph for the maritime domain automatically constructed from natural-language data collected from two main sources: (a) the Worldwide Threats To Shipping (WWTTS)\footnote{\url{https://msi.nga.mil/Piracy}} and (b) the Regional Cooperation Agreement on Combating Piracy and Armed Robbery against Ships in Asia (ReCAAP)\footnote{\url{https://www.recaap.org/}}. We extract the relevant entities and concepts as well as their semantic relations, together with the uncertainty associated with the extracted knowledge. We consider the extracted maritime knowledge graph as our main database for building a QA system. Our training and test corpora include 1,235 and 217 piracy reports in the database, respectively.

\subsection{Maritime Semantic Parsing Dataset}
With 31 grammar rules, the list of variables and the Maritime database, we automatically synthesize 341,381 canonical utterances paired with SQL queries. As discussed above, there are semantic and grammatical mismatches between the synthetic canonical examples and real-world user-issued ones. Paraphrasing the synthesized dataset can potentially generate linguistically more diverse questions. Below, we test the paraphrasing strategies described in Section~\ref{sec:approach} to improve diversity of utterances while ensuring quality which yields the best end-to-end performance.
\subsubsection{Paraphrased Question Collection}
\cite{oren-etal-2021-finding} shows that, with a proper sampling method, we can also achieve decent performance with much less synthetic training data. Following this insight, we only select a subset of synthetic canonical examples (10\%) to reduce our paraphrase and training cost. We use a similar approach to Uniform Abstract Template Sampling (UAT) in~\cite{oren-etal-2021-finding} to get the diversified SQL logical forms structures for paraphrasing.
There are 35,050 canonical examples in our sampled dataset, with 49\% of samples containing abstract variables. 
\subsubsection{SQL Query Writing and Validation}
We evaluate our semantic parser with real-world crowdsourced questions. Specifically, we collect 231 simple and complex questions about a set of given piracy reports. 
For all collected questions, we write their SQL queries manually and randomly divided them into a validation set with 76 examples and a test with 154 examples.

\subsection{Paraphrase Generation}
We employ and evaluate paraphrasing techniques with the following settings. 


\paragraph{Back-translation} We choose three languages in Google Translation: Chinese (medium-resource), Spanish (high-resource) and Telugu (low-resource). The back-translated generation took approximately one day to complete the sampled data.
In the initial paraphrased utterances, we found that Google Translation failed to distinguish the variable names in natural languages, which required the post-processing to correct these variable names. For example, variables like $\$pos$ will be translated to $\$ POS$ or be removed in the paraphrased utterances. Another observation was that the performance of back-translated augmentation highly depends on the language, as shown in Table \ref{tab:filtering}. 
\paragraph{Prompt-based language model generation}
We use (GPT-3\_Davinci) which is the most powerful model in the GPT-3 family. For GPT-3, generation quality can be boosted when more detailed and instructive prompts are given. Due to the limited data we have, we instruct GPT-3 to generate paraphrased questions in the zero-shot manner. Specifically, GPT-3 will directly generate relevant context by giving our manually-designed textual prompts. Different prompts will result in different outputs. The prompts and some generated examples can be found in Table \ref{tab:gpt3_prompt}. The GPT-3 API from OpenAI generates paraphrase randomly based on the seed, and thus may have limited control on the generated output.
\begin{table}[]
\caption{Different prompts result in different paraphrase generation. The paraphrased sentence in \textcolor{cyan}{blue} is our expected output, while sentences in \textcolor{purple}{purple} are not in the ideal format.}

Prompt: I'm a professional and creative paraphraser. I'm required to paraphrase the Original sentence in other words. The output sentence must not be the same as the given sentence. \\

Original: How many heavy lift vessels have been approached  in \$loc ? \\
Paraphrased: \textcolor{cyan}{How many heavy lift vessels have been contacted in \$loc?} \\

\noindent\rule{.5\textwidth}{1pt}\\
\\
Prompt: I'm a professional and creative paraphraser.\\

Original: How many heavy lift vessels have been approached  in \$loc ? \\
Paraphrased: \textcolor{purple}{How many heavy lift vessels have been approached  in \$loc ?} \\
\noindent\rule{.5\textwidth}{1pt}\\
\\
Prompt: I'm a professional and creative paraphraser. I'm required to paraphrase the Original sentence in other words. \\

Original: How many heavy lift vessels have been approached  in \$loc ? \\
Paraphrased: \textcolor{purple}{How many heavy lift vessels have been contacted in your area?} \\
\label{tab:gpt3_prompt}
\vspace{- 0.1 cm}
\end{table}
\begin{figure*}[t!]
    \centering
    \includegraphics[width=0.9\textwidth]{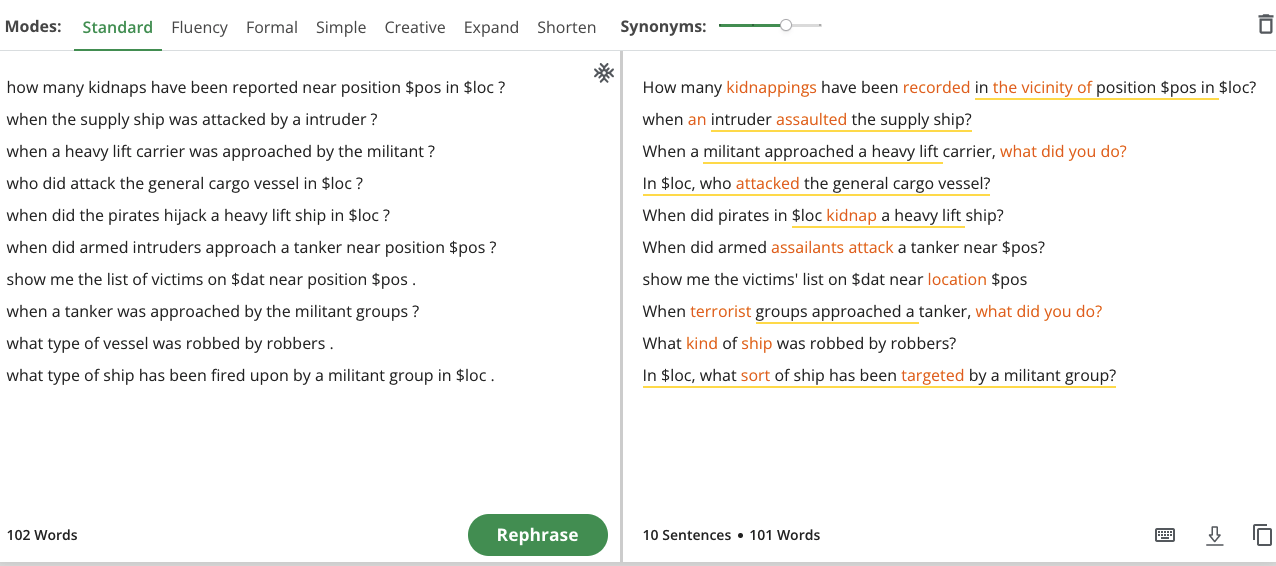}
    \caption{We set the Quillbot paraphraser to the standard mode and the synonyms to have more changes.}
    \label{fig:quillbot}
\end{figure*} 
\paragraph{Fine-tuned autoregressive language model generation}
We employ a pre-trained language model, (BART\_Large) ~\cite{lewis2020bart}, to rewrite a canonical utterance {\it {u}} to a more diverse alternative.
The BART\_Large model\footnote{We use the official implementation in fairseq, https://github.com/pytorch/fairseq.} is fine-tuned on a corpus of high-quality paraphrases sub-sampled from PARANMT \cite{wieting2018paranmt} released by \cite{krishna2020reformulating}.
The model therefore learns to produce paraphrases with a variety of linguistic patterns, which is essential when paraphrasing from canonical utterances.
The generated paraphrases are noisy or potentially vague. We reject paraphrases for which the parser cannot predict their SQL queries (Section~\ref{sec:filtering}).

\paragraph{Commercial paraphrasing model}
As described in the previous section, we employ the commercial paraphrasing tool Quillbot to generate high-quality paraphases. 
The setting of the paraphraser interface can be found in Figure \ref{fig:quillbot}.



\begin{table*}[t!]
\caption{\% of examples kept after each filtering round. The lowest \% in each round is highlighted in \textcolor{cyan}{blue}. The highest \% in each round is highlighted in \textcolor{purple}{purple}.}
\vspace{- 0.1 cm}
\centering
\resizebox{.85\textwidth}{!}{
\begin{tabular}{l|llllll}
\hline
\textbf{\#Rounds}  & BT (Spanish) & BT (Telugu) &  BT (Chinese) & GPT-3 & BART & Quillbot  \\ \hline
Round 1 &  25.15 &  18.52 &\textcolor{cyan}{6.67} &\textcolor{purple}{44.96} & 20.22 & 42.20   \\
Round 2 &    29.63    &   37.85    &  \textcolor{cyan}{13.78}  & 51.81    &   22.34    &    \textcolor{purple}{51.91}     \\
Round 3 &  31.23 &  38.56   &  \textcolor{cyan}{15.42}  &53.22 & 23.35 & \textcolor{purple}{54.25}   \\
\hline
Total &  30.21 &  40.91   &  \textcolor{cyan}{16.37}  & 53.37 & 19.68 & \textcolor{purple}{54.92}   \\
\hline

\end{tabular}
}
\centering
\label{tab:filtering}
\end{table*}
\subsection{Paraphrase Filtering}
\label{sec:filtering}
We develop a scheme where we input the paraphrased question into a parser, which in turn generates the SQL queries. If the SQL query matches exactly with the one generated from the original question of the paraphrase, we consider that this question has been correctly paraphrased. 


Table \ref{tab:filtering} shows the filtering results of a number of paraphrasing techniques, including back-translation, fine-tuned BART, GPT-3 and Quillbot. In general, we see that the GPT-3 from OpenAI and Quillbot significantly outperform the other approaches. In contrast, the fine-tuned BART achieves only 20.22\% retention. Moreover, we observe that GPT-3 has the highest semantic preserving ability by keeping 44.96\% of the paraphrased sentences in the first round of filtering and Quillbot by keeps 51.91\% and 54.25\% in the second and third rounds of filtering suggesting prompt-based and commercial methods have good potential in paraphrasing.
The Back-Translation method on Chinese performs the worst. We conjecture that Chinese is far from English in terms of linguistic structure; thus, Back-Translation suffers from translation errors. With more rounds of re-training, the generalization ability of the parser is improved, so more examples are kept. However, the improvement becomes marginal after the second round of re-training. Therefore, we only adopt three rounds since re-training is time-consuming.

\subsection{Paraphrase Generation Method Analysis}
\begin{table}[t!]
\caption{Practical Performance (Time comparison per 1000 samples). \\ \textbf{*}The measurement of BART was run on a single RTX 8000 NVIDIA GPU.}
\vspace{- 0.1 cm}
\centering
\begin{tabular}{l|cccc|c}
\hline 
\textbf{Techniques} & BT & GPT-3 & \textbf{*}BART & Quillbot & All \\ \hline
Time (minutes) & 8.67  & 33.3 & 0.43 & 60 & 102.4\\
\hline
\end{tabular}

\label{tab:practical}

\end{table}
\begin{table*}[t!]
\caption{Diversity Evaluation on Paraphrase Methods. The lowest score in each BLEU metric is highlighted in \textcolor{cyan}{blue}. The highest score in each BLEU metric is highlighted in \textcolor{purple}{purple}.}
\vspace{-0.1 cm}
\centering
\resizebox{.85\textwidth}{!}{
\begin{tabular}{l|llllll}
\hline
\textbf{Metrics}  & BT (Spanish) & BT (Telugu) &  BT (Chinese) & GPT-3 & BART & Quillbot  \\ \hline
BLEU-1 & 59.05  & 48.97 & \textcolor{cyan}{36.95} & \textcolor{purple}{74.81} & 53.39 & 59.74    \\
BLEU-2 & 43.65  & 36.26 & \textcolor{cyan}{23.89} & \textcolor{purple}{68.42} & 40.32 & 51.22    \\
BLEU-3 & 33.22  & 26.60 & \textcolor{cyan}{15.98} & \textcolor{purple}{63.48} & 31.12 & 44.63   \\
BLEU-4 & 25.63  & 19.33 & \textcolor{cyan}{10.30} & \textcolor{purple}{59.46} & 24.17 & 39.55   \\
\hline

\end{tabular}
}
\centering
\label{tab:diversity}
\end{table*}
\textbf{Paraphrase diversity} We measure the diversity of each paraphrase method by using the  BLEU metric \cite{papineni-etal-2002-bleu}. BLEU-$N$ was originally proposed for automatic machine translation evaluation, which calculates the similarity score based on $N$-grams. The lower the score is, the more dissimilar the generated context is to the reference. The dissimilarity can imply the diversity of the paraphrased context when used in paraphrase evaluation. GPT-3 paraphrasing method generates the least diversified sentences, while Back-Translation (Chinese) generates the most diversified ones. Quillbot and BART can generate paraphrases with a close diversity level. However, more Quillbot paraphrases are kept, indicating that Quillbot can perform well in terms of both diversity and the preservation of semantics. An interesting finding is that the results in Table \ref{tab:diversity} are relatively consistent with the filtering report in Table \ref{tab:filtering}. We assume it is easier for the parser to generalize on the paraphrases which have more words overlapping with the original questions.   

\textbf{Time-efficiency}
Table \ref{tab:practical} shows the required time for paraphrasing 1000 canonical utterances by each technique. The pre-trained BART model is the fastest way of paraphrasing while Quillbot is the slowest one. 


\textbf{Controllability.} Among various paraphrasing techniques, Quillbot provides different settings and various paraphrased candidates (Figure \ref{fig:quillbot}). Current neural models still find it more challenging to get all possible replaced words or phrases while maintaining the semantic meaning.

\textbf{Open-source.} The second most promising paraphrasing model right now is provided by Quillbot. No other open-source model, except GPT-3, can achieve excellent quality compared to Quillbot. What techniques are practical for paraphrasing needs further investigation.

\textbf{Compute resource.} Unlike back-translation, where we can adopt translation systems to the paraphrasing tasks, direct paraphrasing needs specific paraphrasing data to train a good model. Designing a good paraphrasing dataset across different domains is challenging.

\textbf{Readability.} The back-translation system requires increasing human readability. Some translated sentences are not human-readable. A method should be proposed to automatically self-reject some invalid results. This will aid the back translation based paraphrasing method by providing more valid data.

\subsection{Semantic Parser Evaluation}
\begin{table*}[]
\caption{Evaluation of the semantic parser baselines based on additional paraphrased data using different techniques. }
\vspace{- 0.1 cm}
\centering
\resizebox{1\textwidth}{!}{
\begin{tabular}{l|ccc|ccc}
\hline
\multirow{2}{*}{\textbf{Training Data}} & \multicolumn{3}{c|}{\textbf{SEQ2SEQ}}       & \multicolumn{3}{c}{\textbf{RoBERTa-base}}  \\ \cline{2-7} 
                                        & \textbf{\begin{tabular}[c]{@{}c@{}}Exact match\\ (acc)\end{tabular}} & \textbf{\begin{tabular}[c]{@{}c@{}}Exact match\\ no order (acc)\end{tabular}} & \textbf{\begin{tabular}[c]{@{}c@{}}Compo\_match\\  (F1)\end{tabular}} & \textbf{\begin{tabular}[c]{@{}c@{}}Exact match\\  (acc)\end{tabular}} & \textbf{\begin{tabular}[c]{@{}c@{}}Exact match\\ no order (acc)\end{tabular}} & \textbf{\begin{tabular}[c]{@{}c@{}}Compo\_match\\  (F1)\end{tabular}} \\ \hline
Original dataset  & 36.77  & 48.39  & 70.71 & 43.07   & 57.79 & 77.26 \\
BT (Spanish)  & 37.42 & 50.97  & 75.09  & 42.86  & 57.14 & 77.11 \\
BT (Chinese) & 36.77  & 54.84  & 74.61  & 45.44  & 59.44 & 78.17 \\
BT (Telugu)  & 34.84  & 49.68 & 73.19 & 42.86 & 57.14 & 76.72  \\
BART  & 36.77  & 54.84 & 75.60  & 45.02 & 59.93 & 78.93  \\
GPT-3 & \textbf{43.87}  & \textbf{58.86} & \textbf{76.79} & 47.05 & 61.64  & 79.04 \\
Quilbot & 41.29 & 55.48 & 75.53  & \textbf{50.65} & \textbf{62.99}  & \textbf{80.58} \\
Full Filtered Data  & 40.00  & 56.13 & 76.19 & 46.75 & 60.17 & 78.55 \\ \hline
\end{tabular}
}
\label{tab:experiment_1}
\centering
\end{table*}
\textbf{Baseline Models.}
We utilize two common methods of semantic parsers for the evaluation of different data settings: i) the attentional \seqseq\cite{luong2015effective} framework which uses LSTMs~\cite{hochreiter1997long} as the encoder and decoder, and ii) \bertlstm\cite{xu2020schema2qa}, a Seq2Seq framework with a copy mechanism~\cite{gu2016incorporating} which uses RoBERTa~\cite{liu2019roberta} as the encoder and LSTM as the decoder. The inputs for these models are the natural-language questions and the outputs are their sequence of linearized SQL queries' tokens. Our evaluation metrics include Exact Matching as in~\cite{dong2018coarse,li2021few}, Exact Matching (no-order), and Component Matching as in~\cite{yu2018spider}

\textbf{Component Matching:} To conduct a detailed
analysis of model performance, we measure the
average exact match we are reporting F1 between the prediction and ground truth on different SQL components. For each of the following components: {\fontfamily{qcr}\selectfont • SELECT • FROM • WHERE • GROUP BY • ORDER BY}.
We decompose each component in the prediction and the ground truth as bags of several sub-components, and check whether or not these two sets of components match exactly. In our evaluation,
we treat each component as a set so that for example, {\fontfamily{qcr}\selectfont WHERE va.aggressor = "pirates" AND va.victim = "container ship"} and {\fontfamily{qcr}\selectfont WHERE va.victim = "container ship" AND va.aggressor = "pirates"} would be treated as the same query.

\textbf{Exact Matching:} We measure whether the predicted query as a whole is equivalent to the ground truth query. 

\textbf{Exact Matching (no-order):} We first evaluate the SQL clauses and ignore the order in each component. The predicted query is correct only if all of the components are correct. 

Table \ref{tab:experiment_1} shows the performance of parsers on different training datasets. We have merged the filtered data using various paraphrasing techniques with the original training dataset and trained the semantic parsers with the new training dataset. We observe that Back-Translation paraphrase hurts the semantic parsing, indicating the generated sentences are too diversified and considered to be noisy signal during training. Among these Back-Translation methods, the model trained on filtered Back-Translation paraphrases with low-resource Telugu performs the worst. This may result from limited training data when training Google Translation on Telugu. We also find that paraphrasing only 10\% of the original training dataset using GPT-3 and Quillbot can constantly boost the semantic parsing.
We further demonstrate in Table \ref{tab:qb_paraphrasing} that with more Quillbot paraphrase data, the performance of semantic parser keeps increasing. With the findings in Tables \ref{tab:filtering} and \ref{tab:diversity}, we conclude paraphrase filtering and corresponding diversity are not directly related to the semantic parsing performance. A more comprehensive and efficient approach needs to be found to evaluate paraphrase quality in terms of semantic parsing.

\begin{table}[]
\caption{The effect of data paraphrasing portion on filtering and parser (RoBERTa-base) logical form matching accuracy}
\vspace{-0.1 cm}
\resizebox{0.5\textwidth}{!}{
\begin{tabular}{l|cc}
\hline
\textbf{Training   Data}  & \multicolumn{1}{c}{\textbf{Filtering}} & \multicolumn{1}{c}{\textbf{Compo\_match (F1)}} \\ \hline
Original training dataset &  -  &  77.26 \\
With Quillbot (10\%) & 54.25   &  80.58  \\
With Quillbot (20\%) & 64.73  & 81.73   \\ \hline
\end{tabular}
}
\label{tab:qb_paraphrasing}
\end{table}

\section{Observations and Discussions}
By conducting several experiments and  using the obtained results on semantic parsers , we now present our observations and insights on the task.

\begin{itemize}
\item With the finding in Table \ref{tab:experiment_1} we observe that paraphrasing is a powerful technique to improve the performance of the final QA system without additional human efforts.

\item Table \ref{tab:qb_paraphrasing} shows that with only 10\% additional paraphrased questions (using Quillbot) we are able to increase the average F1 by 3.32 percent points and with further 10\% additional paraphrased questions, F1 increases by 4.47 percent points. We can conclude that a small portion of high quality paraphrased data would be sufficient for training semantic parsers. 

\item Among languages we choose for Back-Translation paraphrasing in TABLE \ref{tab:diversity}, Chinese shows the lowest BLEU score and the highest diverse paraphrased questions. Subsequently, the semantic parser (RoBERTa-base) demonstrates a better performance with Chinese compare to the other two languages (i.e. Spanish and Telugu). 
Further analyses is required to investigate the impact of amount of filtering data and diversity on parsers' performance.


\item In the last row of TABLE \ref{tab:experiment_1}, we simply combined the all paraphrased data from all paraphrasing techniques. In the future, we will investigate whether combining the data from only more effective paraphrasing techniques, such as Quillbot and GPT-3, can induce more promising results.
\end{itemize}

\section{Conclusion }
This work is aimed at enhancing human-machine communications by automatically tranlating users' questions expressed in natural language into SQL queries. By developing a compact set of synchronous grammar rules enhanced with a set of  multiple automatic paraphrasing and filtering techniques in order to generate large scale training data, we are able to train a semantic parser without the need to use any (manually) annotated training dataset. The experimential results demonstrated  promising results could be achieved at a very low cost, which is important in many real-world situations where data is scarce and annotated data is not available.

\bibliographystyle{plain}

\bibliography{IEEEabrv}

\end{document}